\pdfoutput=1
\documentclass[11pt]{article}

\usepackage{acl}
\usepackage{times}
\usepackage{latexsym}
\usepackage[T1]{fontenc}
\usepackage[utf8]{inputenc}
\usepackage{microtype}
\usepackage{inconsolata}
\usepackage{graphicx}
\usepackage{subcaption}
\usepackage{booktabs}
\usepackage{multirow}
\usepackage{makecell}
\usepackage{amsmath}
\usepackage{amssymb}
\usepackage{amsthm}
\usepackage{xcolor}
\usepackage{cleveref}
\usepackage{algorithm}
\usepackage{algpseudocode}
\usepackage{enumitem}
\usepackage{xspace}

\newcommand{\maybeincludegraphics}[2][]{%
  \IfFileExists{figures/#2}{%
    \includegraphics[#1]{figures/#2}%
  }{%
    \fbox{\parbox[c][0.22\textheight][c]{0.95\linewidth}{\centering\texttt{#2}}}%
  }%
}

\newcommand{\method}{\textsc{FCPRAG}\xspace}
\newtheorem{proposition}{Proposition}

\definecolor{mycolor_1}{RGB}{0, 0, 175}
\definecolor{mycolor_2}{RGB}{190, 60, 0}

\usepackage{listings}
\usepackage[most]{tcolorbox}
\usepackage{needspace}
\tcbuselibrary{listings,breakable,skins}

\lstdefinelanguage{udiff}{
  morecomment=[f][\color{red!70!black}]{-},
  morecomment=[f][\color{green!50!black}]{+},
  morecomment=[f][\color{blue!60!black}]{@@},
  morecomment=[f][\color{gray!70}]{---},
  morecomment=[f][\color{gray!70}]{+++},
}

\lstdefinestyle{promptdiff}{
  language=udiff,
  basicstyle=\ttfamily\scriptsize,
  columns=fullflexible,
  keepspaces=true,
  showstringspaces=false,
  breaklines=true,
  breakatwhitespace=false,
  tabsize=2,
}

\newtcblisting{PromptDiffBoxGray}[2][]{%
  enhanced,
  breakable,
  colback=black!2,
  colframe=black!35,
  boxrule=0.35pt,
  arc=1.2mm,
  left=1.2mm,right=1.2mm,top=0.8mm,bottom=0.8mm,
  fonttitle=\bfseries\footnotesize,
  title={#2},
  listing only,
  listing options={
  style=promptdiff,
  escapeinside={(*@}{@*)}
  },
  #1
}

\newtcolorbox{PromptDiffBoxBlue}[2][]{%
  enhanced,
  breakable,
  colback=cyan!4!white,
  colframe=cyan!28!white,
  colbacktitle=cyan!20!white,
  coltitle=black,
  boxrule=0.35pt,
  arc=1.2mm,
  left=1.2mm,right=1.2mm,top=0.8mm,bottom=0.8mm,
  fonttitle=\bfseries\footnotesize,
  title={#2},
  #1
}

\newtcblisting{PromptDiffBoxRed}[2][]{%
  enhanced,
  breakable,
  colback=red!5,
  colframe=red!35,
  boxrule=0.35pt,
  arc=1.2mm,
  left=1.2mm,right=1.2mm,top=0.8mm,bottom=0.8mm,
  fonttitle=\bfseries\footnotesize,
  title={#2},
  listing only,
  listing options={
  style=promptdiff,
  escapeinside={(*@}{@*)}
  },
  #1
}
\definecolor{darkred}{RGB}{0, 0, 0}
\newcommand{\mlyang}[1]{\textcolor{darkred}{#1}}

\definecolor{darkblue}{RGB}{0, 0, 139}

\definecolor{darkgreen}{RGB}{0, 100, 0}

\newtcblisting{PromptDiffBoxGreen}[2][]{%
  enhanced,
  breakable,
  colback=green!5,
  colframe=green!80,
  boxrule=0.35pt,
  arc=1.2mm,
  left=1.2mm,right=1.2mm,top=0.8mm,bottom=0.8mm,
  fonttitle=\bfseries\footnotesize,
  title={#2},
  listing only,
  listing options={
  style=promptdiff,
  escapeinside={(*@}{@*)}
  },
  #1
}

\newtcblisting{PromptDiffBoxPink}[2][]{%
  enhanced,
  breakable,
  colback=pink!5,
  colframe=pink!80,
  boxrule=0.35pt,
  arc=1.2mm,
  left=1.2mm,right=1.2mm,top=0.8mm,bottom=0.8mm,
  fonttitle=\bfseries\footnotesize,
  title={#2},
  listing only,
  listing options={
  style=promptdiff,
  escapeinside={(*@}{@*)}
  },
  #1
}

\newtcblisting{PromptDiffBoxOrange}[2][]{%
  enhanced,
  breakable,
  colback=orange!5,
  colframe=orange!50,
  boxrule=0.35pt,
  arc=1.2mm,
  left=1.2mm,right=1.2mm,top=0.8mm,bottom=0.8mm,
  fonttitle=\bfseries\footnotesize,
  title={#2},
  listing only,
  listing options={
  style=promptdiff,
  escapeinside={(*@}{@*)}
  },
  #1
}
\title{FCPRAG: Fusion-Controller Parametric Retrieval-Augmented Generation for Stable Multi-Passage LoRA Injection}
\author{
  \textbf{Jinchang Zhu$^{1}$, Jindong Li$^{1}$, YI DING$^{1}$, Nie Xiaojian$^{1}$, Rong Fu$^{2}$,} \\
  \textbf{Shuangyong Song$^{3}$, Haowei He$^{3}$, Menglin Yang$^{1,\dagger}$} \\
  $^{1}$The Hong Kong University of Science and Technology (Guangzhou) \\
  $^{2}$University of Macau \\
  $^{3}$Institute of Artificial Intelligence (TeleAI), China Telecom \\
  \texttt{jzhu997@connect.hkust-gz.edu.cn \quad menglinyang@hkust-gz.edu.cn}
}

\begin{document}
\maketitle
\begingroup
\renewcommand{\thefootnote}{\fnsymbol{footnote}}
\footnotetext[2]{Corresponding author.}
\endgroup
\begin{abstract}
Parametric retrieval-augmented generation (PRAG) injects retrieved evidence into a large language model (LLM) through passage-specific LoRA adapters, reducing reliance on long in-context prompts.
When multiple passages are retrieved for the same query, however, \emph{evidence-level fusion} becomes a bottleneck: equal-weight merging can amplify weak or conflicting evidence, and translating retrieval signals into fusion weights often requires fragile global tuning.
We propose \textbf{\method{}}, a fusion-controlled parametric RAG framework that adds a lightweight controller for retrieval-conditioned, sample-level adapter fusion.
The controller predicts per-passage fusion scores together with sample-level calibration signals, including a mixing gate and an adaptive temperature, enabling fusion that stays selective under informative retrieval signals and conservative under uncertainty.
\method{} is trained with merge-aware supervision derived from each adapter's marginal contribution within a multi-adapter merge, using training data only.
We further show that a single dataset-level temperature is suboptimal under heteroscedastic retrieval uncertainty, motivating sample-level adaptation.
Experiments on HotpotQA, 2WikiMultiHopQA, PopQA, and ComplexWebQuestions (CWQ) across three LLM backbones show that \method{} consistently improves F1 over standard RAG and parametric RAG baselines, with gains of up to 4.65\% on 2WikiMultiHopQA and 7.55\% on CWQ, while also reducing tuning cost and improving robustness under retrieval perturbations.
\end{abstract}

\section{Introduction}

Retrieval-augmented generation (RAG) improves the factuality of large language models (LLMs) by retrieving external evidence and conditioning generation on the retrieved content~\cite{rag2020,dpr2020,fid2021}.
However, classical in-context RAG increases prompt length with the number and size of retrieved passages, leading to higher inference cost and potential degradation under long or conflicting contexts.
Recent RAG research therefore explores stronger retrieval and generation control to reduce redundancy and error propagation~\cite{active_rag2023,selfrag2024}, but the fundamental trade-off remains: multi-evidence reasoning often benefits from more context, while longer contexts are expensive and fragile.

Parametric RAG (PRAG)~\cite{prag2025} offers an alternative: instead of appending passages to the prompt, it \emph{parameterizes} each passage into lightweight LoRA weights~\cite{lora2022} during an offline stage, and loads these passage-specific parameters at test time to inject knowledge into the model.
This design reduces online context length and admits an optional combine setting that also appends retrieved passages to the input prompt (Section~\ref{sec:prag_baseline}). 
Recent PRAG-family methods further reduce cost by using translators or routing to reuse a small set of latent experts~\cite{dyprag2025,polyprag2025}.

While these methods successfully reduce inference cost by moving evidence from the prompt into model parameters, they introduce a new challenge: \emph{multi-evidence fusion}.
For many knowledge-intensive queries, retrieval returns multiple relevant passages (often partially relevant, redundant, or mutually inconsistent).
In PRAG-family systems, this yields multiple passage-specific LoRA experts that must be merged before generation.
Uniform fusion (equal weights) can inject noise from weak passages, while hand-tuned fusion heuristics (e.g., globally set softmax temperature or blending factors) often fail to generalize across queries, tasks, and model scales.
This sensitivity affects both efficiency and robustness: in PRAG-style parametric injection, the fusion weights directly control which evidence is written into the model's effective parameters at test time, so even small calibration mismatches can over- or under-inject passage updates (Figure~\ref{fig:calibration_ab}(a)).
As a result, PRAG-family pipelines frequently rely on expensive grid search over global temperature and blending coefficients to reach strong performance.

To understand why this grid search is difficult to eliminate, we identify three properties that make this fusion bottleneck particularly challenging in parametric RAG.
\textbf{First, evidence utility is highly sample-dependent.}
Even under a fixed retriever, the same query can surface a mix of strongly supportive, weakly relevant, redundant, or distracting passages, and the \emph{marginal contribution} of each passage-specific adapter (estimated by removing one adapter from the merge and measuring the F1 drop) can vary substantially across instances and subtasks.
Uniform fusion therefore tends to either dilute a dominant adapter (when a single passage is decisive) or over-inject weak adapters (when retrieval is noisy).
For example, in bridge-style multi-hop QA, one retrieved passage may contain the bridge entity needed to answer the question, while others are distractors.
\textbf{Second, parametric injection amplifies fusion mistakes.}
Unlike in-context conditioning, where the model may ignore or down-weight unhelpful tokens at generation time, parametric RAG fuses evidence through weight-space updates: an over-weighted passage adapter directly alters the model's effective parameters for the current query, making conflicts and noise harder to ``wash out'' via prompting.
\textbf{Third, mapping scores to weights is itself a calibration problem.}
Even if a scoring model ranks passages reasonably well, the optimal sharpness of the resulting distribution (temperature) and the degree of deviation from uniform fusion (blending) can change across backbones, tasks, and retrieval quality, which is why grid search is often needed to stabilize performance.
This motivates learning \emph{per-sample} calibration so that fusion can be selective when confident and conservative when uncertain, while avoiding costly per-model tuning (Figures~\ref{fig:calibration_ab} and~\ref{fig:tuning_cost}).


\begin{table}[t]
\centering
\caption{Capability comparison across RAG, PRAG-family systems, and FCPRAG.}
\resizebox{\linewidth}{!}{%
{\renewcommand{\arraystretch}{1.4}%
\begin{tabular}{lcccc}
\toprule
\textbf{Method} &
\makecell{\textbf{Parametric}\\\textbf{evidence injection}} &
\makecell{\textbf{Fusion in}\\\textbf{parameter space}} &
\makecell{\textbf{Sample-level}\\\textbf{calibrated fusion}} &
\makecell{\textbf{Uniform-prior}\\\textbf{fallback}} \\
\midrule
RAG~\cite{rag2020}       & \texttimes & \texttimes & \texttimes & \texttimes \\
PRAG~\cite{prag2025}      & \checkmark & \checkmark & \texttimes & \texttimes \\
DyPRAG~\cite{dyprag2025}    & \checkmark & \checkmark & \texttimes & \texttimes \\
Poly-PRAG~\cite{polyprag2025} & \checkmark & \checkmark & \texttimes & \texttimes \\
\textbf{\method (Ours)}    & \checkmark & \checkmark & \checkmark & \checkmark \\
\bottomrule
\end{tabular}}%
}
\label{tab:capability_comparison}
\end{table}

At a high level, this fusion problem resembles model and adapter merging~\cite{model_soups2022,task_arithmetic2022,ties_merging2023,adapterfusion2021,ainsworth2022git,jin2022dataless,stoica2023zipit,izmailov2018averaging,matena2022merging}.
However, PRAG-family fusion is fundamentally \emph{retrieval-conditioned}: the adapters to be merged are tied to passages retrieved for a specific query, and the appropriate confidence in a peaked vs.\ conservative weighting can change from sample to sample.
This motivates treating fusion as a learned, calibrated decision rather than a fixed global heuristic, analogous in spirit to classical calibration of predictive distributions~\cite{guo2017calibration,platt1999probabilistic,niculescu2005predicting,kuleshov2018accurate}.

To solve the above challenges, we propose \method{} which introduces a learned \emph{Fusion Controller} into the PRAG inference pipeline.
The controller predicts sample-level fusion weights for the retrieved passages and jointly learns per-sample calibration signals to stabilize the mapping from raw fusion scores to final merge weights.
Importantly, \method{} is designed as a plug-in: it does not change the underlying PRAG offline parameterization, retrieval, or generation, and can replace the merge step in PRAG/DyPRAG with minimal engineering.
We empirically validate that this learned calibration can replace manual hyperparameter tuning (Figure~\ref{fig:tuning_cost}).


\begin{figure*}[t]
  \centering
  \includegraphics[width=\textwidth]{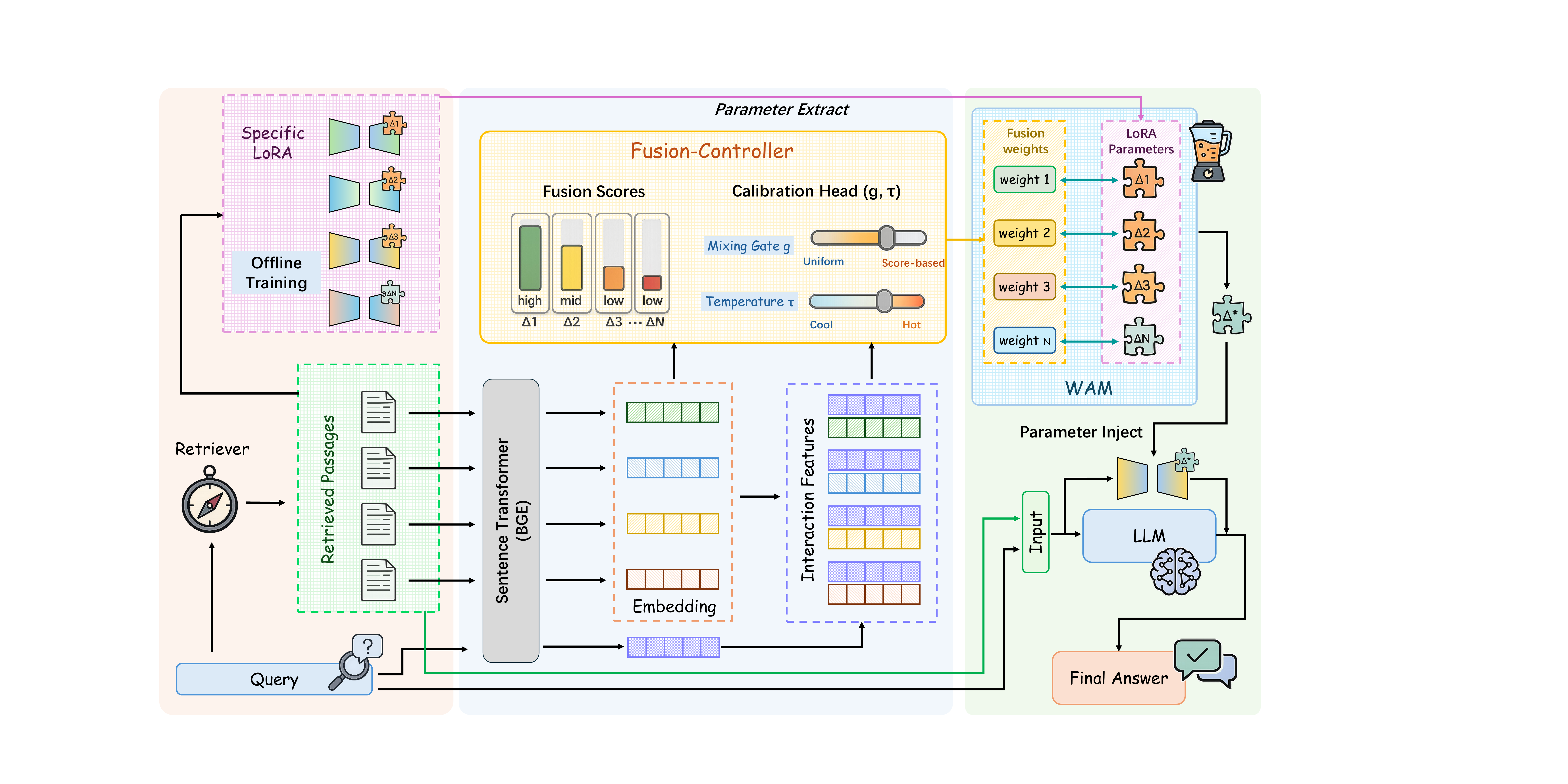}
  \caption{Overview of \method{}. Given a query, retrieve multiple passages and load their passage-specific LoRA adapters (offline-trained). A Fusion Controller predicts per-passage weights and per-sample calibration, then a weighted adapter merger (WAM) merges the adapters for parametric injection (optionally combined with in-context passages).}
  \label{fig:pipeline}
\end{figure*}

\textbf{Why fusion control is the right interface.}
A key shift in parametric RAG is that evidence is injected through \emph{parameters} rather than \emph{tokens}.
When retrieval returns $N$ passages, PRAG-family systems produce $N$ passage-LoRA adapters that must be fused into a single injected update.
This fusion step becomes the main place where evidence relevance, redundancy, and conflict must be resolved.
Our design therefore targets the fusion interface directly: we learn (i) evidence-level weights and (ii) per-sample calibration, so the system can be both selective when confident and conservative when uncertain. To summarize, in this study, we made the following contributions.
\begin{itemize}
\item \textbf{\method{} architecture.} We introduce a learned Fusion Controller for PRAG-family inference that performs retrieval-conditioned, sample-level LoRA fusion and calibration, adding $<1\%$ trainable parameters relative to the backbone LLM. 
  \item \mlyang{\textbf{Theoretical motivation.} We prove (Proposition~\ref{prop:main}) that under heteroscedastic retrieval uncertainty, any single dataset-level fusion temperature incurs strictly positive calibration regret, formally motivating sample-level $(g,\tau)$ adaptation.}
\item \textbf{Merge-aware supervision under no-leak.} We propose a training-signal construction that supervises fusion weights using each passage adapter's marginal contribution within a multi-LoRA merge, while enforcing a no-leak train/eval separation. 
\item \textbf{Empirical gains and key ablations.} On HotpotQA~\cite{hotpotqa2018} and 2WikiMultiHopQA~\cite{ho2020multi}, \method{} improves F1 across three LLM backbones (Llama-3.2-\allowbreak 1B~\cite{llama2023}, Qwen2.5-\allowbreak 1.5B~\cite{2025_arXiv_Qwen2.5_Qwen2.5-Technical-Report}, Llama-3-\allowbreak 8B~\cite{llama2023}) and achieves the best Avg.\ Rank in Table~\ref{tab:main}. A random controller ablation is substantially worse, validating the learned fusion mechanism. 
\end{itemize}

\section{Related Work}
Unlike in-context RAG, which injects retrieved evidence by concatenating passages to the prompt, parametric RAG injects retrieved knowledge as parameter updates~\cite{rag2020,dpr2020,fid2021,guu2020retrieval,borgeaud2022improving,izacard2022few,prag2025}. This avoids long-context overhead, but under multi-passage retrieval it introduces a different online bottleneck: multiple evidence-conditioned updates must be combined before generation.

PRAG~\cite{prag2025} converts each retrieved passage into passage-specific LoRA weights via offline fine-tuning~\cite{lora2022,houlsby2019parameter,li2021prefix,han2024parameter}. DyPRAG~\cite{dyprag2025} reduces per-passage training and storage by dynamically generating parametric knowledge at inference time, while Poly-PRAG~\cite{polyprag2025} routes queries to a small set of latent LoRA experts. Despite these differences in offline parameterization, PRAG-family systems share the same online fusion problem: when several passages are retrieved, the model must merge multiple passage-conditioned parametric updates, often under noisy or partially conflicting evidence.

Existing fusion methods mainly operate either in token/prediction space, such as Fusion-in-Decoder and RePlug~\cite{fid2021,shi2024replug}, or in task-level parameter composition, such as model soups, task arithmetic, TIES-Merging, and AdapterFusion~\cite{model_soups2022,task_arithmetic2022,ties_merging2023,adapterfusion2021}. Our setting is different: we study \emph{retrieval-conditioned evidence-level fusion in parameter space}, where useful merge weights vary across queries and over-weighting weak evidence can inject spurious knowledge. \method{} addresses this gap by learning both per-query fusion scores and calibration that controls how confidently the model departs from uniform fusion. Extended background is provided in Appendix~\ref{app:rw_extended}.

\section{Methodology}
\label{sec:method}

\subsection{Problem Formulation and PRAG Baseline}
\label{sec:prag_baseline}
We consider knowledge-intensive QA where, given a query $q$, a retriever returns a set of passages $\mathcal{P}=\{p_1,\dots,p_N\}$.
PRAG~\cite{prag2025} parameterizes each passage $p_i$ into a lightweight passage-LoRA adapter $\Delta_i$ by offline fine-tuning on synthetic QA-style supervision derived from $p_i$.
At test time, PRAG loads the offline-trained adapters $\{\Delta_i\}_{i=1}^{N}$ corresponding to the retrieved passages $\mathcal{P}$ and merges them into a single adapter $\Delta(\mathcal{P})$, which is injected into the LLM to generate an answer.
We denote this parametric-only inference as \textbf{Parametric-only} (\textbf{Param.}), and an optional variant \textbf{Combine} where the same retrieved passages are also provided to the LLM in-context in addition to parametric injection (following the ``PRAG + in-context'' combination described in~\cite{prag2025}).

\paragraph{Data augmentation (training-side).}
Following the PRAG-family pipeline, we construct synthetic supervision for each passage $p_i$ using an instruction-tuned LLM: (i) a rewritten variant of $p_i$ and (ii) a small set of QA pairs whose answers are supported by $p_i$.
These augmentation artifacts define a passage-specific training set $D_i$.

\paragraph{Passage-LoRA training (offline parameterization).}
We then fine-tune a passage-specific LoRA adapter $\Delta_i$ on QA-style prompts built from $D_i$, optimizing the standard causal language modeling loss while keeping the backbone LLM frozen.
To encourage parametric encoding rather than relying only on in-context evidence, training mixes \emph{passage-conditioned} examples (the prompt includes $p_i$ or its rewrite) and \emph{question-only} examples (the prompt omits the passage but still supervises the answer).
Appendix~\ref{app:train_prompts} lists the training-side prompt templates used for augmentation and passage-LoRA fine-tuning.

\subsection{Evidence-level Multi-LoRA Fusion Setting}
\label{sec:multi_lora_setting}
In the multi-passage setting, each passage yields a passage-LoRA adapter $\Delta_i$ and the system must fuse $\{\Delta_i\}_{i=1}^N$ before generation.
Uniform fusion (equal weights) treats all passages as equally useful, which can be suboptimal when passages have different relevance or when some passages introduce conflicting knowledge.
We aim to learn a \emph{fusion policy} that assigns weights $w_i(q,\mathcal{P})$ per query so that the merged adapter emphasizes helpful evidence and suppresses noise.

\subsection{\method{}: Fusion Controller with Learned Calibration}
\label{sec:fcprag_controller}
Figure~\ref{fig:pipeline} shows the high-level pipeline.
\method{} augments PRAG-family inference with a lightweight Fusion Controller $f_\theta$.
Given a query and its retrieved passages, the controller predicts retrieval-conditioned fusion scores for the corresponding passage-LoRA adapters, and outputs per-sample calibration signals to stabilize the mapping from scores to weights.
We denote the downstream weighted merge block as a \emph{Weighted Adapter Merger} (WAM), which applies controller weights to retrieved adapters and outputs one merged adapter for injection.
Inference follows a simple sequence: encode the query and retrieved passages, predict fusion scores and sample-wise calibration, map them to merge weights, merge the retrieved adapters, and generate in either Parametric-only or Combine mode. Appendix~\ref{app:inference_pseudocode} gives pseudocode.

\paragraph{Inputs.}
For controller inputs, we construct interaction features from dense query and passage embeddings produced by a sentence embedding model.
Let $\mathbf{e}_q \in \mathbb{R}^{d}$ be the query embedding and $\mathbf{E}_p = [\mathbf{e}_{p_1},\dots,\mathbf{e}_{p_N}] \in \mathbb{R}^{N\times d}$ be passage embeddings (normalized).
We form interaction features for each passage:
\begin{equation}
  \mathbf{x}_i = \left[\mathbf{e}_q;\ \mathbf{e}_{p_i};\ \mathbf{e}_q \odot \mathbf{e}_{p_i};\ |\mathbf{e}_q-\mathbf{e}_{p_i}|\right].
  \label{eq:interaction}
\end{equation}

\paragraph{Outputs and weight mapping.}
The controller predicts a scalar score $s_i \in (0,1)$ for each passage via an MLP, and a calibration head predicts:
(i) a gate $g \in (0,1)$ and (ii) a temperature $\tau \in [\tau_{\min},\tau_{\max}]$.
We map scores to a probability distribution and blend with a uniform prior:
\begin{align}
  \pi_i &= \mathrm{softmax}\left(\frac{s_i}{\tau}\right), \\
  w_i &= g \cdot \pi_i + (1-g)\cdot \frac{1}{N}. \label{eq:weights}
\end{align}
Intuitively, $\tau$ controls how sharply scores are converted into weights, while $g$ controls how far the controller deviates from uniform fusion.
This mirrors the theoretical role of calibration below: heteroscedastic retrieval uncertainty motivates sample-level sharpness, and the uniform-prior interpolation in Eq.~\ref{eq:weights} implements shrinkage toward conservative fusion when score signals are unreliable.
The theory supports this calibration choice rather than claiming optimality of the full controller architecture.

\paragraph{Weighted Adapter Merger (WAM).}
Given retrieved adapters $\{\Delta_i\}_{i=1}^{N}$ and controller outputs $\{w_i\}_{i=1}^{N}$ from Eq.~\ref{eq:weights}, we first apply merge-time normalization:
\begin{equation}
  \tilde{w}_i = \frac{N\,w_i}{\sum_{j=1}^{N} w_j + \epsilon}.
\end{equation}
WAM then applies the normalized weights as scaling factors and composes the scaled adapters into a single merged adapter:
\begin{equation}
  \Delta^\star = \mathcal{M}_{\mathrm{WAM}}\!\left(\{\tilde{w}_i \cdot \Delta_i\}_{i=1}^{N}\right).
\end{equation}
Finally, we inject $\Delta^\star$ into the frozen backbone model and denote the adapted model by $\mathcal{M}_{\Delta^\star}$:
\begin{equation}
  \begin{aligned}
  \mathcal{M}_{\Delta^\star} &\gets \textsc{Inject}(\mathcal{M}, \Delta^\star),\\
  \hat{a} &\gets \textsc{Generate}(\mathcal{M}_{\Delta^\star}, \text{prompt}).
  \end{aligned}
\end{equation}

\subsection{Merge-aware Training Signal and No-leak Protocol}
\label{sec:labels}
Training the Fusion Controller requires supervision that reflects how each passage-LoRA adapter contributes under \emph{multi-LoRA fusion} (as opposed to isolated single-adapter behavior).
We therefore construct \emph{merge-aware} marginal contribution labels using training-side data only:
for each training instance $(q,\mathcal{P},a)$, we evaluate the system with all $N$ adapters merged under uniform fusion to obtain $F_1^{\text{all}}$, then remove the $i$-th adapter and re-evaluate to obtain $F_1^{\setminus i}$; the difference measures the marginal effect of that adapter under fusion.
We define
\begin{equation}
  \delta_i = F_1^{\text{all}} - F_1^{\setminus i},
\end{equation}
where we use $\delta_i$ to denote a scalar marginal contribution (to avoid confusion with LoRA deltas $\Delta_i$).
We convert $\{\delta_i\}$ into a soft target distribution
\begin{equation}
  y_i = \mathrm{softmax}\left(\frac{\delta_i}{T_{\text{label}}}\right).
\end{equation}
We then train the controller to match this target distribution via KL divergence (below); when contributions are flat (no preference), the target becomes uniform and we down-weight these samples to prevent learning a trivial uniform mapping.
A schematic illustration and label-source analysis are deferred to Appendix~\ref{app:label_analysis}.
Label construction and controller training use training data only, while evaluation is performed on a held-out split.

\paragraph{Objective.}
Let $\mathbf{w}_\theta(q,\mathcal{P})$ be the controller's predicted distribution from Eq.~\ref{eq:weights}.
For sample $b$, we define
\begin{align}
  \ell_{\text{KL}}^{(b)}
  &= \sum_{i=1}^{N} y_i^{(b)} \log \frac{y_i^{(b)}}{w_i^{(b)} + \epsilon},
\end{align}
and optimize the weighted batch objective
\begin{align}
  \mathcal{L}_{\text{KL}}
  &= \frac{\sum_b \omega^{(b)} \ell_{\text{KL}}^{(b)}}{\sum_b \omega^{(b)} + \epsilon},
\end{align}
where $\omega^{(b)}$ is the sample weight.
When the deltas are flat (no preference), $\mathbf{y}$ becomes uniform; we down-weight these samples to prevent the model from learning a trivial uniform mapping.


\begin{table*}[!t]
\centering
\caption{Main results (F1, \%) on 2WikiMultiHopQA, HotpotQA, PopQA, and ComplexWebQuestions (CWQ). Suffix ``-Combine'' denotes results under the Combine setting. All baseline methods (Vanilla, Standard RAG, PRAG, DyPRAG) were manually reproduced under the same experimental setup for fair comparison. \textbf{Avg. Rank} is the mean rank (lower is better) over the 8 non-Avg columns within each backbone block. Best results are highlighted in \textcolor{mycolor_1}{\textbf{bold}} and second-best results are \underline{underlined}.}
\label{tab:main}
\setlength{\tabcolsep}{4pt}
\renewcommand{\arraystretch}{1.03}
\resizebox{\textwidth}{!}{%
\begin{tabular}{@{}llcccccccccc@{}}
    \toprule
    \multirow{2}{*}{\textbf{Base LLM}} & \multirow{2}{*}{\textbf{Method}} & \multicolumn{5}{c}{\textbf{2WikiMultiHopQA}} & \multicolumn{2}{c}{\textbf{HotpotQA}} & \multirow{2}{*}{\textbf{PopQA}} & \multirow{2}{*}{\textbf{CWQ}} & \multirow{2}{*}{\makecell[c]{\textbf{Avg.}\\\textbf{Rank}}} \\
    \cmidrule(lr){3-7} \cmidrule(lr){8-9}
    & & \textbf{Compare} & \textbf{Bridge} & \textbf{Inference} & \textbf{Compose} & \textbf{Avg.} & \textbf{Bridge} & \textbf{Compare} &  &  &  \\
    \midrule
    \multirow{8}{*}{\makecell[c]{\textbf{Llama-3.2-1B}}}
    & \textbf{Vanilla} & 36.48 & 34.23 & 13.80 & 5.60 & 22.53 & 9.85 & 37.84 & 14.56 & 28.52 & 8.00 \\
    & \textbf{Standard RAG} & 41.22 & 39.31 & 15.57 & 7.22 & 25.83 & 11.42 & 39.39 & 17.15 & 31.78 & 6.38 \\
    & \textbf{PRAG} & \textcolor{mycolor_1}{\textbf{46.83}} & \textcolor{mycolor_1}{\textbf{44.32}} & 17.16 & 8.55 & \underline{29.22} & 12.99 & 41.84 & 19.16 & 32.09 & 3.62 \\
    & \textbf{PRAG-Combine} & 41.03 & 41.26 & 19.37 & 8.21 & 27.47 & \underline{21.35} & 42.73 & \underline{32.94} & 31.82 & 4.31 \\
    & \textbf{DyPRAG} & 41.53 & 37.98 & 18.67 & 8.21 & 26.60 & 10.84 & 42.91 & 18.69 & 31.92 & 5.31 \\
    & \textbf{DyPRAG-Combine} & \underline{44.87} & 41.42 & 19.31 & \underline{9.42} & 28.76 & 11.35 & 45.67 & 21.14 & 33.87 & 3.25 \\
    \cmidrule(lr){2-12}
    & \textbf{\method{} (Our)} & 44.29 & \underline{42.14} & \underline{20.21} & 5.97 & 28.15 & 14.56 & \textcolor{mycolor_1}{\textbf{47.49}} & 17.28 & \textcolor{mycolor_1}{\textbf{34.93}} & \underline{3.12} \\
    & \textbf{\method{}-Combine} & 44.11 & 41.47 & \textcolor{mycolor_1}{\textbf{21.67}} & \textcolor{mycolor_1}{\textbf{10.65}} & \textcolor{mycolor_1}{\textbf{29.48}} & \textcolor{mycolor_1}{\textbf{23.05}} & \underline{45.60} & \textcolor{mycolor_1}{\textbf{38.33}} & \underline{34.17} & \textcolor{mycolor_1}{\textbf{2.00}} \\
    \midrule

    \multirow{8}{*}{\makecell[c]{\textbf{Qwen2.5-1.5B}}}
    & \textbf{Vanilla} & 39.53 & 37.26 & 15.20 & 6.40 & 24.60 & 13.43 & 43.23 & 16.58 & 31.25 & 6.62 \\
    & \textbf{Standard RAG} & 40.89 & 33.12 & 18.79 & 11.10 & 25.97 & 16.32 & 43.96 & 19.73 & 31.50 & 4.88 \\
    & \textbf{PRAG} & 43.85 & 35.41 & \textcolor{mycolor_1}{\textbf{20.17}} & \underline{12.77} & \underline{28.05} & 17.35 & 46.22 & 18.90 & \underline{32.68} & 3.62 \\
    & \textbf{PRAG-Combine} & 40.62 & 37.95 & 18.03 & 7.58 & 26.04 & \underline{18.71} & 39.33 & \underline{23.01} & 21.48 & 5.12 \\
    & \textbf{DyPRAG} & \underline{43.91} & 36.54 & 17.98 & 9.12 & 26.89 & 11.69 & 43.18 & 19.92 & 31.54 & 5.25 \\
    & \textbf{DyPRAG-Combine} & 42.23 & 40.83 & 18.56 & 9.75 & 27.84 & 12.13 & \underline{46.41} & \textcolor{mycolor_1}{\textbf{32.08}} & 32.29 & \underline{3.50} \\
    \cmidrule(lr){2-12}
    & \textbf{\method{} (Our)} & \textcolor{mycolor_1}{\textbf{45.59}} & \underline{42.89} & 17.81 & \textcolor{mycolor_1}{\textbf{13.18}} & \textcolor{mycolor_1}{\textbf{29.87}} & 18.58 & \textcolor{mycolor_1}{\textbf{51.54}} & 21.30 & \textcolor{mycolor_1}{\textbf{35.27}} & \textcolor{mycolor_1}{\textbf{2.38}} \\
    & \textbf{\method{}-Combine} & 38.97 & \textcolor{mycolor_1}{\textbf{43.73}} & \underline{19.04} & 7.78 & 27.38 & \textcolor{mycolor_1}{\textbf{19.98}} & 40.80 & 20.77 & 20.02 & 4.62 \\
    \midrule

    \multirow{8}{*}{\makecell[c]{\textbf{Llama-3-8B}}}
    & \textbf{Vanilla} & 50.88 & 48.60 & 21.49 & 12.68 & 33.41 & 37.80 & 61.00 & 36.39 & 39.88 & 7.25 \\
    & \textbf{Standard RAG} & 55.46 & 53.68 & 20.94 & 11.80 & 35.47 & 41.25 & 63.43 & 37.97 & 41.29 & 6.50 \\
    & \textbf{PRAG} & 58.11 & 56.29 & 21.34 & 12.73 & 37.12 & 43.49 & 66.03 & 39.06 & 43.08 & 5.00 \\
    & \textbf{PRAG-Combine} & 62.47 & \underline{58.23} & \underline{29.94} & \underline{18.92} & \underline{42.39} & \underline{75.18} & \textcolor{mycolor_1}{\textbf{82.83}} & \underline{42.29} & 42.49 & \underline{2.38} \\
    & \textbf{DyPRAG} & 55.32 & 52.87 & 20.44 & 11.92 & 35.14 & 43.76 & 62.92 & 35.87 & 41.14 & 7.00 \\
    & \textbf{DyPRAG-Combine} & 59.81 & 56.94 & 27.65 & 18.82 & 40.81 & 71.67 & 81.12 & 39.21 & 42.96 & 3.38 \\
    \cmidrule(lr){2-12}
    & \textbf{\method{} (Our)} & \textcolor{mycolor_1}{\textbf{67.07}} & 56.53 & 25.88 & 16.16 & 41.41 & 43.95 & 70.64 & 41.33 & \textcolor{mycolor_1}{\textbf{50.63}} & 3.12 \\
    & \textbf{\method{}-Combine} & \underline{67.04} & \textcolor{mycolor_1}{\textbf{62.39}} & \textcolor{mycolor_1}{\textbf{32.86}} & \textcolor{mycolor_1}{\textbf{25.88}} & \textcolor{mycolor_1}{\textbf{47.04}} & \textcolor{mycolor_1}{\textbf{76.15}} & \underline{81.87} & \textcolor{mycolor_1}{\textbf{43.79}} & \underline{48.73} & \textcolor{mycolor_1}{\textbf{1.38}} \\
    \bottomrule
\end{tabular}%
}
\end{table*}

\subsection{Plug-and-play Integration}
\label{sec:plugin}
\method{} is designed to be a drop-in fusion module:
\textbf{PRAG} uses uniform weights in the merge; \method{} replaces those weights with $\mathbf{w}_\theta$.
\textbf{DyPRAG} additionally uses translator-generated deltas; \method{} can be applied by replacing mean pooling/averaging in the aggregation stage with a weighted sum driven by $\mathbf{w}_\theta$.
\textbf{Poly-PRAG} performs routing over latent experts; we discuss how a calibrated fusion controller could complement routing decisions in future work, but do not claim full plug-and-play integration.

\subsection{Efficiency and Overhead}
\label{sec:efficiency}
The Fusion Controller is a small MLP operating on retriever embeddings.
Compared to PRAG-family inference, the additional overhead is a single embedding forward pass for the query and passages plus lightweight MLP computation.
Controller training is fast and does not require backpropagating through the LLM; label construction is the main offline cost but can be parallelized and reused across controller variants.

\subsection{Theoretical Motivation}
\label{sec:theory}

We provide a formal justification for sample-level fusion calibration.
The full proof is in Appendix~\ref{app:proofs}.

\begin{proposition}[Limitation of Dataset-level Fusion Calibration]\label{prop:main}
Under heteroscedastic retrieval uncertainty, different queries induce different optimal fusion sharpness levels.
Consequently, no single dataset-level temperature can be optimal for all queries, and any fixed global temperature incurs positive expected calibration regret under the risk-aware objective analyzed in Appendix~\ref{app:proofs}:
\begin{equation}\label{eq:regret-def}
    \mathrm{Reg}(T)
    \;\triangleq\; \mathbb{E}_{t,\,\mathbf{s}}\!\Big[
        \mathcal{R}_t\!\bigl(\mathbf{w}(T_t^\star;\,\mathbf{s})\bigr)
        - \mathcal{R}_t\!\bigl(\mathbf{w}(T;\,\mathbf{s})\bigr)
    \Big],
\end{equation}
In particular, even the best dataset-level temperature
\begin{equation}\label{eq:global-opt}
    T_{\mathrm{global}}^\star
    \;\in\; \arg\max_{T>0}\;\mathbb{E}_{t,\,\mathbf{s}}\!\Big[\mathcal{R}_t\!\bigl(\mathbf{w}(T;\,\mathbf{s})\bigr)\Big]
\end{equation}
satisfies $\mathrm{Reg}(T_{\mathrm{global}}^\star) > 0$.
The full assumptions and proof are deferred to Appendix~\ref{app:proofs}.
\end{proposition}

\noindent\textbf{Intuition.}
Softmax temperature controls how sharply evidence is fused in parameter space.
Under low retrieval uncertainty, sharp fusion is beneficial; under high uncertainty, conservative (near-uniform) fusion is safer.
When uncertainty varies across queries, a single dataset-level temperature cannot satisfy both regimes.
This directly motivates the calibration head in Eq.~\ref{eq:weights}: $\tau$ adapts fusion sharpness, while $g$ allows the learned distribution to shrink back toward the uniform prior under unreliable retrieval signals.
Appendix~\ref{app:oracle} derives the corresponding oracle-weight form.

\section{Experimental Setup}
\label{sec:setup}
We evaluate on HotpotQA~\cite{hotpotqa2018}, 2WikiMultiHopQA~\cite{ho2020multi}, PopQA~\cite{mallen2023not}, and ComplexWebQuestions (CWQ)~\cite{talmor2018web}, using three instruction-tuned backbones: Llama-3.2-1B, Qwen2.5-1.5B, and Llama-3-8B~\cite{llama2023,2025_arXiv_Qwen2.5_Qwen2.5-Technical-Report}. Following PRAG-family evaluation practice~\cite{prag2025,dyprag2025}, we use \textbf{F1} (\%) as the primary metric. For HotpotQA and 2WikiMultiHopQA, we report subtask-level F1; for 2Wiki we also report macro average over its four subtasks.

\label{sec:setup_baselines}
We compare Vanilla, Standard RAG~\cite{rag2020,dpr2020,fid2021}, PRAG~\cite{prag2025}, DyPRAG~\cite{dyprag2025}, and \method{}. All methods use the same retriever and the same top-$K$ passages per query; only the knowledge injection and fusion mechanism differ. For parametric methods (PRAG, DyPRAG, and \method{}), we report both Parametric-only and Combine settings (Section~\ref{sec:prag_baseline}). A randomly initialized controller is evaluated only as an ablation (Table~\ref{tab:2wiki-ablation-random}). Full implementation details are deferred to Appendix~\ref{app:implementation}. Runtime and implementation overhead are discussed in Appendix~\ref{app:implementation}.

\section{Experiments}
\label{sec:experiments}

\begin{table}[t]
\centering
\caption{Ablation on 2WikiMultiHopQA (F1 scores, Qwen2.5-1.5B and Llama-3-8B). We compare uniform fusion (PRAG), a random (untrained) fusion controller, and our learned fusion controller (\method{}) under Parametric-only and Combine settings.}
\label{tab:2wiki-ablation-random}
\setlength{\tabcolsep}{3.5pt}
\renewcommand{\arraystretch}{1.03}
\resizebox{\columnwidth}{!}{%
\begin{tabular}{@{}llccccc@{}}
\toprule
\textbf{Base LLM} & \textbf{Method} & \textbf{Cmp.} & \textbf{Br.} & \textbf{Inf.} & \textbf{Compst.} & \textbf{Avg.} \\
\midrule
\multirow{6}{*}{\rotatebox[origin=c]{90}{\textbf{Qwen2.5-1.5B}}}
 & PRAG & 43.85 & 35.41 & \textbf{20.17} & 12.77 & 28.05 \\
 & PRAG-Combine & 40.62 & 37.95 & 18.03 & 7.58 & 26.04 \\
\cmidrule(lr){2-7}
 & Random Ctrl. & 44.39 & 36.05 & 16.10 & 7.95 & 26.12 \\
 & Random Ctrl.-Combine & 38.98 & 41.87 & 18.32 & 7.32 & 26.62 \\
\cmidrule(lr){2-7}
 & \method{} (Ours) & \textbf{45.59} & 42.89 & 17.81 & \textbf{13.18} & \textbf{29.87} \\
 & \method{}-Combine & 38.97 & \textbf{43.73} & 19.04 & 7.78 & 27.38 \\
\midrule
\multirow{6}{*}{\rotatebox[origin=c]{90}{\textbf{Llama-3-8B}}}
 & PRAG & 58.11 & 56.29 & 21.34 & 12.73 & 37.12 \\
 & PRAG-Combine & 62.47 & 58.23 & 29.94 & 18.92 & 42.39 \\
\cmidrule(lr){2-7}
 & Random Ctrl. & 60.18 & 54.12 & 22.05 & 12.11 & 37.12 \\
 & Random Ctrl.-Combine & 64.58 & 59.01 & 28.47 & 17.18 & 42.31 \\
\cmidrule(lr){2-7}
 & \method{} (Ours) & \textbf{67.07} & 56.53 & 25.88 & 16.16 & 41.41 \\
 & \method{}-Combine & 67.04 & \textbf{62.39} & \textbf{32.86} & \textbf{25.88} & \textbf{47.04} \\
\bottomrule
\end{tabular}%
}
\end{table}

\subsection{Main Results}
Table~\ref{tab:main} reports results on 2WikiMultiHopQA, HotpotQA, PopQA, and ComplexWebQuestions (CWQ).
Across backbones and benchmarks, \method{} generally improves over PRAG's uniform merge, with gains most pronounced on multi-evidence settings where retrieved passages vary in usefulness or contain distractors.
These results support our thesis that retrieval-conditioned evidence fusion is a first-class bottleneck in parametric RAG, and that learning both weights and calibration provides a practical upgrade over globally tuned heuristics.
A varying-$K$ check is reported in Appendix~\ref{app:varying_k}: performance is not monotonic in the Parametric-only setting, but \method{}-Combine improves over PRAG-Combine at $K=4$ and $K=5$.

\subsection{Ablation: Random Fusion Controller}
To verify that gains come from \emph{learned} fusion rather than architectural side effects, we replace the Fusion Controller with a random (untrained) one while keeping the same parametric-fusion pipeline.
Table~\ref{tab:2wiki-ablation-random} (Qwen2.5-1.5B and Llama-3-8B) shows that the random controller underperforms \method{} across subtasks under both Parametric-only and Combine settings.
This indicates that improvements are not explained by adding an extra module or changing the merge operator; they come from learning retrieval-conditioned fusion weights (and calibration) that map evidence to meaningful fusion decisions.

\Needspace{7\baselineskip}
\paragraph{Merge-aware supervision.}
The controller also depends on how fusion labels are constructed.
Holding the scoring projector fixed on HotpotQA (Qwen2.5-1.5B, Parametric-only), replacing isolated single-adapter labels with our merge-aware labels improves macro F1 from 32.47 to 33.91, with the largest gain on comparison questions (+2.63).
This links the leave-one-out construction in Section~\ref{sec:labels} to downstream fusion quality; Appendix~\ref{app:label_analysis} reports the full label-source ablation, attribution diagnostics, and offline cost.

\subsection{Analysis: Where Does Fusion Control Help?}
The four 2Wiki subtasks stress different reasoning patterns and evidence usage.
In bridge/comparison questions, a single passage can dominate the answer, and uniform fusion may over-inject irrelevant passage-specific updates.
In compositional/inference questions, multiple pieces of evidence may be required, and the system must balance contributions without amplifying conflicts.
\method{} addresses both regimes by (i) learning to emphasize dominant experts when confident (high $g$ and low $\tau$) and (ii) falling back toward uniform fusion when signals are ambiguous (low $g$), improving robustness.

\begin{figure}[t]
  \centering
  \maybeincludegraphics[width=\linewidth]{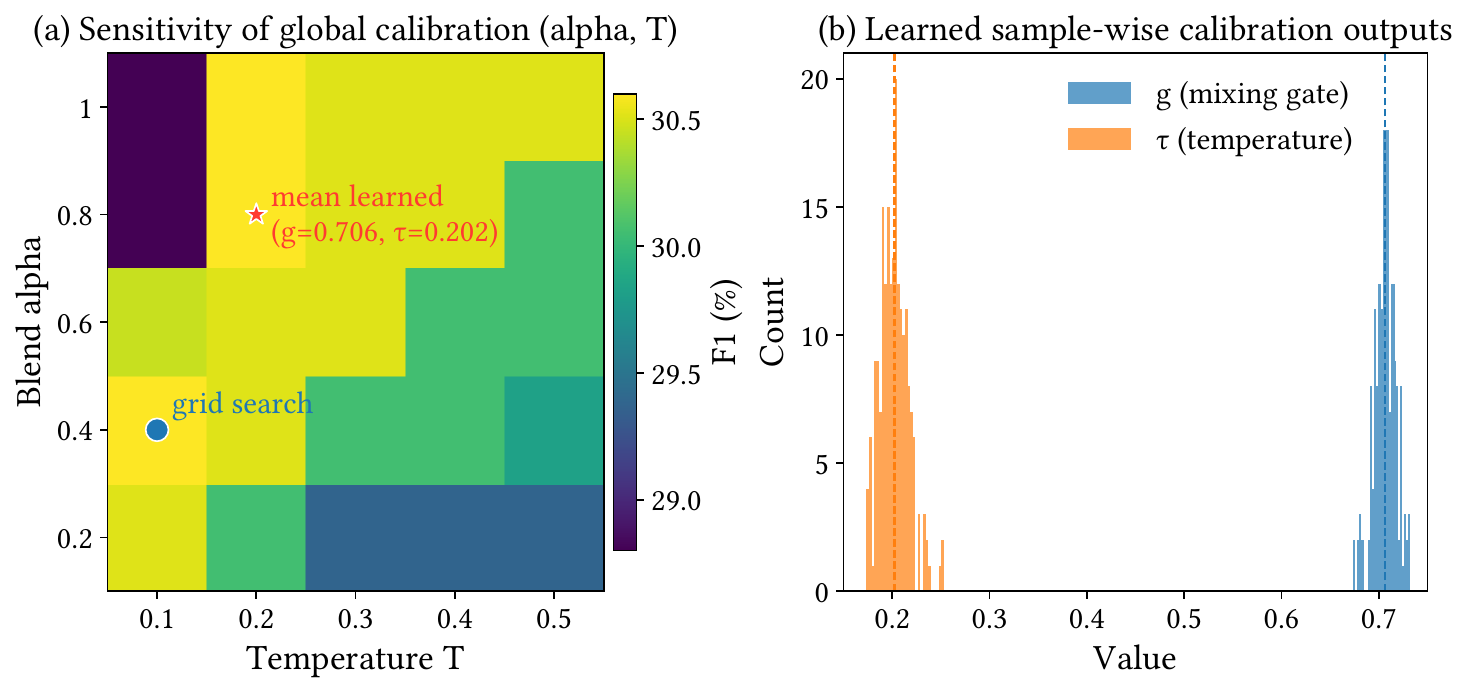}
  \caption{Calibration diagnostics on HotpotQA. \textbf{(a)} F1 over global blend $\alpha$ and temperature $T$ for a score-only fusion baseline. \textbf{(b)} Learned sample-wise calibration outputs $(g,\tau)$ from \method{}.}
  \label{fig:calibration_ab}
\end{figure}

\begin{figure}[t]
  \centering
  \maybeincludegraphics[width=\linewidth]{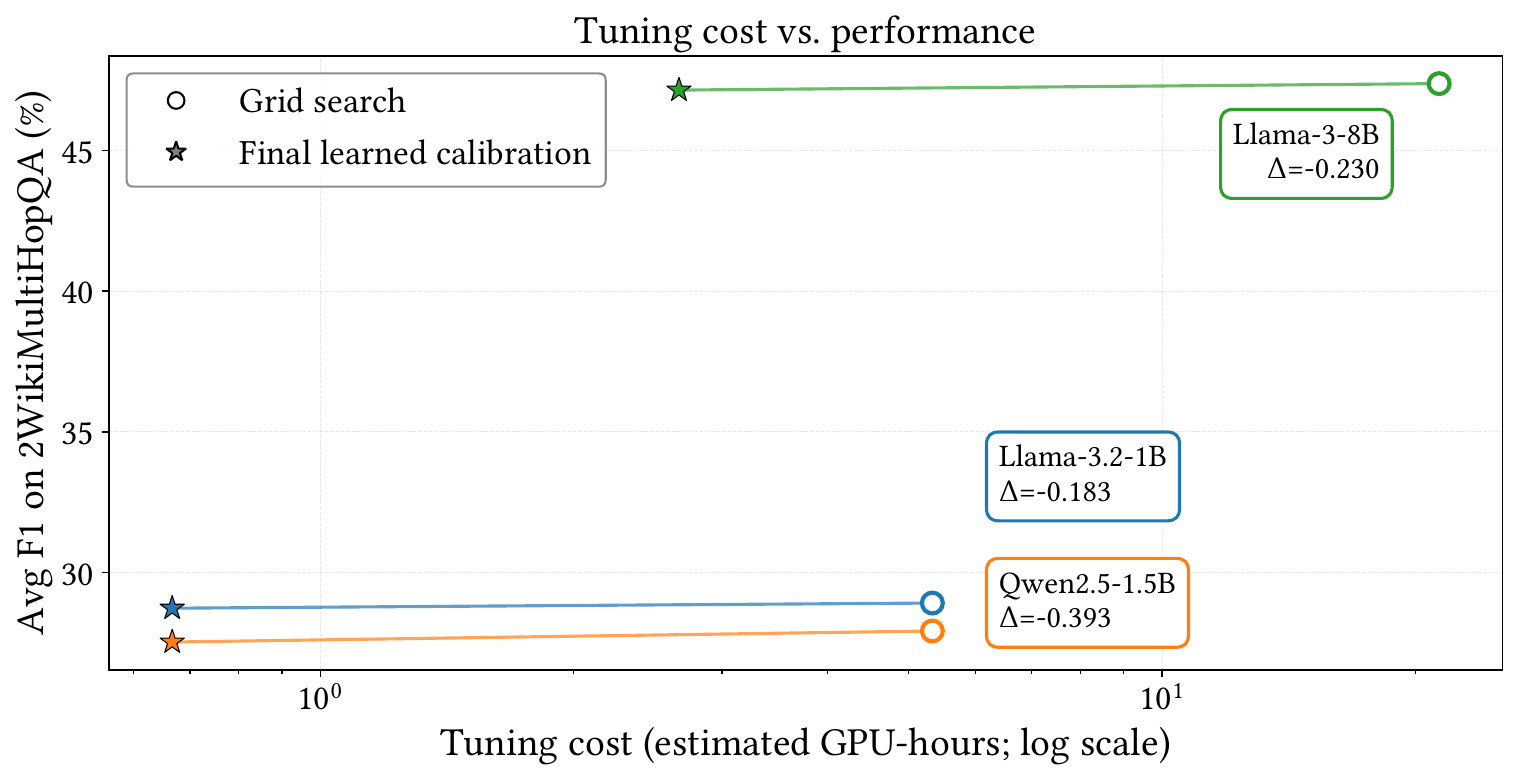}
\caption{Tuning cost vs.\ performance on 2WikiMultiHopQA (Combine): each point is macro-average F1 over the four 2Wiki subtasks. Circles denote grid-selected global calibration; stars denote \method{}'s learned calibration.}
  \label{fig:tuning_cost}
\end{figure}

\paragraph{Calibration replaces grid search.}
Figures~\ref{fig:calibration_ab}--\ref{fig:tuning_cost} show that learned calibration can replace per-model global tuning.
The score-only global baseline is sensitive to blend/temperature choices, whereas \method{} predicts sample-wise $(g,\tau)$ that avoid degenerate pure-uniform or overly sharp regimes and remain in a stable high-performing basin.
This is consistent with Proposition~\ref{prop:main}: fixed dataset-level calibration is brittle under heteroscedastic retrieval uncertainty.
Although the learned mean $(g,\tau)$ does not exactly match the single best global $(\alpha,T)$ cell, it stays in the same high-performance region, yielding strong F1 without exhaustive tuning.
On 2Wiki in the \emph{Combine} setting, learned calibration achieves comparable performance to grid-selected settings with substantially lower tuning cost, without changing the retriever, merge operator, or offline passage parameterization.

\subsection{Robustness to Retrieval Perturbations}
\label{sec:robust}
\begin{figure}[t]
  \centering
  \maybeincludegraphics[width=\linewidth]{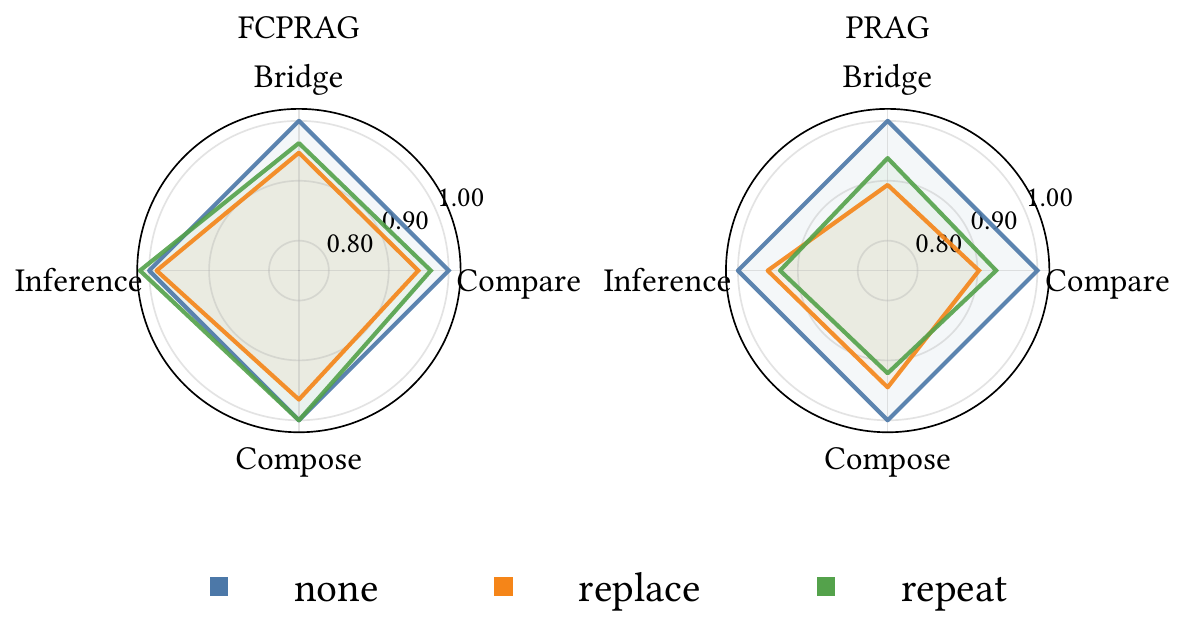}
  \caption{Retrieval perturbation robustness on 2WikiMultiHopQA (Llama-3-8B, Parametric-only; $K=3$). We report \textbf{replace-one} and \textbf{repeat-one} perturbations, with each panel normalized by the method's own unperturbed score (none $=1$).}
  \label{fig:retrieval_perturb}
\end{figure}

Real retrieval systems can return irrelevant or redundant passages, which is especially harmful when evidence is injected through adapter weights.
We test retrieval noise at $K=3$ with two one-passage perturbations: \textbf{replace-one} substitutes an irrelevant passage, and \textbf{repeat-one} duplicates a retrieved passage.
Figure~\ref{fig:retrieval_perturb} shows that \method{} retains more relative performance than PRAG under both perturbations, consistent with conservative fusion when retrieval signals are unreliable.
Because each panel is normalized by the method's own unperturbed score, the figure emphasizes relative performance retention rather than absolute score differences.
This matches the role of adaptive calibration: noisier retrieved sets should shift fusion toward less aggressive injection instead of applying a fixed global temperature.

\section{Conclusion}
\label{sec:conclusion}

We presented \method{}, a learned fusion controller for stable, retrieval-conditioned multi-passage LoRA fusion in parametric retrieval augmented generation.
By predicting per-passage weights and per-sample calibration, \method{} improves robustness and reduces the need for expensive global calibration.
Across four datasets, 2WikiMultiHopQA, HotpotQA, PopQA, and ComplexWebQuestions(CWQ) and three LLM backbones, \method{} improves F1 scores over the baseline methods in most settings, while a random controller performs significantly worse.
These results suggest that sample-level calibration is a practical mechanism for adapting evidence fusion to query-dependent retrieval uncertainty, rather than relying on a single global setting across tasks and backbones.
Because the method operates at the fusion stage, it can strengthen multi-passage parametric RAG with minimal changes to the underlying retrieval, offline parameterization, and generation pipeline.
These results suggest that learned fusion control is a promising and practical direction for parametric RAG systems under multi-passage retrieval and parametric fusion.

\section*{Limitations}
\label{sec:limitations}

\textbf{Offline label cost:} merge-aware supervision requires evaluating one all-merge and $K$ leave-one-out merges per training sample (e.g., four forward passes for $K=3$), increasing offline computation cost and making larger $K$ more expensive.
\textbf{Embedding dependence:} the controller operates on query/passage embeddings; if embedding similarity is misaligned with actual answer utility, fusion predictions may degrade.
\textbf{Scaling to larger evidence sets:} we focus on a small, fixed number of passages per query; extending to larger $K$ may require sparse or hierarchical fusion policies.

\section*{Ethical Considerations}
\label{sec:ethics}

This work studies a retrieval-augmented QA system using public benchmarks and does not involve new human-subject data collection.
The main ethical risk is that parametric injection can make retrieved errors or biased evidence more persistent in the model's effective parameters during inference.
\method{} mitigates this risk by learning to down-weight weak or conflicting evidence, but it does not guarantee factual correctness or fairness.
Deployments should therefore pair the method with source filtering, answer attribution, and task-specific safety evaluation before use in high-stakes settings.

\bibliography{references}

\clearpage
\appendix


\section{Proof of Proposition~\ref{prop:main}}\label{app:proofs}
For a given query $q$, retrieval returns passages $\mathcal{P}=\{p_1,\dots,p_N\}$, each associated with a passage-specific LoRA adapter $\Delta_i$.
A scoring model assigns per-passage scores $\mathbf{s} = (s_1,\dots,s_N)$, which may vary across queries.
We assume scores are non-degenerate, i.e., $\Pr[\exists\, i\neq j: s_i \neq s_j] > 0$; otherwise $\mathbf{w}(T;\,\mathbf{s})$ is uniform for all $T$ and temperature has no effect.
We allow the score distribution $\mathbf{s}\mid t$ to differ across query types.
The proof proceeds by expanding the risk-aware objective, showing that the optimal temperature depends monotonically on the noise variance, and concluding that heteroscedastic types force distinct optima that no single temperature can satisfy.
Fusion produces a merged update
\begin{equation}\label{eq:merge}
    \Delta(T;\,\mathbf{s}) = \sum_{i=1}^{N} w_i(T;\,\mathbf{s})\,\Delta_i,
\end{equation}
where weights are given by a softmax with temperature $T>0$:
\begin{equation}\label{eq:softmax-w}
    w_i(T;\,\mathbf{s}) = \frac{\exp(s_i / T)}{\sum_{j=1}^{N} \exp(s_j / T)}.
\end{equation}
Note that even under a \emph{global} (dataset-level) temperature $T$, the weights $\mathbf{w}(T;\,\mathbf{s})$ are sample-dependent through the scores $\mathbf{s}$.
We show that allowing the temperature to also adapt per sample can strictly reduce the expected calibration regret under heteroscedastic uncertainty.

We model the effective utility of adapter $i$ on a given sample as
\begin{equation}\label{eq:utility}
    U_i = \mu_i + \varepsilon_i,
\end{equation}
where $\mu_i$ is the mean contribution (independent of query type $t$) and $\varepsilon_i$ is zero-mean noise induced by retrieval uncertainty.

\paragraph{Assumptions.}
\begin{enumerate}[label=(A\arabic*),nosep]
    \item \textit{Independence.}
          $\varepsilon_i$ are independent across passages, with
          $\mathrm{Var}(\varepsilon_i \mid \text{type } t) = \sigma_t^2$.
    \item \textit{Non-degenerate heteroscedastic mixture.}
          There exist at least two query types $A,B$ with
          $\sigma_A^2 \neq \sigma_B^2$, each occurring with positive probability
          ($0 < \pi < 1$, where $\pi = \Pr[t\!=\!A]$).
    \item \textit{Score--utility alignment (local).}
          For each type $t$, the expected score--utility interaction
          $\mathbb{E}_{\mathbf{s}\mid t}\!\left[\sum_i \frac{\partial w_i(T;\,\mathbf{s})}{\partial T}\,\mu_i\right]$
          is strictly decreasing in $T$ on an interval containing both $T_A^\star$ and $T_B^\star$;
          intuitively, increasing $T$ shifts mass from higher-utility to lower-utility passages on average.
    \item \textit{Interior optimum.}
          \begin{enumerate}[label=(\alph*),nosep]
              \item For each type $t$, the expected objective
              $\mathbb{E}_{\mathbf{s}\mid t}[\mathcal{R}_t(\mathbf{w}(T;\,\mathbf{s}))]$
              admits a unique maximizer $T_t^\star \in (0,\infty)$.
              \item The derivative term
              $\mathbb{E}_{\mathbf{s}\mid t}\!\left[\sum_i w_i \frac{\partial w_i}{\partial T}\right]$
              is non-zero and does not change sign in a neighborhood of $T_t^\star$.
          \end{enumerate}
\end{enumerate}

\paragraph{Remark on (A1).}
The independence assumption simplifies the variance term;
allowing correlations adds cross-terms $\sum_{i\neq j} w_i w_j \,\mathrm{Cov}(\varepsilon_i,\varepsilon_j)$ but the variance remains type-dependent through $\sigma_t^2$, so the monotonicity argument in~\S\ref{app:uncertainty} still applies and Proposition~\ref{prop:main} holds.
In particular, positive correlations among passage noises can further motivate conservative (near-uniform) fusion under high uncertainty.

We adopt a mean--variance objective as a tractable proxy that captures the key tension between evidence sharpness and noise amplification; the empirical sections validate the conclusion under actual QA metrics.
We define the risk-aware objective for fusion as
\begin{equation}\label{eq:risk-obj}
    \mathcal{R}_t(\mathbf{w})
    = \mathbb{E}\!\left[\sum_i w_i\, U_i\right]
    - \lambda\,\mathrm{Var}\!\left(\sum_i w_i\, U_i\right),
    \quad \lambda > 0.
\end{equation}

The per-type optimal temperature maximizes the expected risk-aware objective:
\begin{equation}\label{eq:type-opt}
    T_t^\star
    \;\in\; \arg\max_{T>0}\;
    \mathbb{E}_{\mathbf{s}\mid t}\!\Big[\mathcal{R}_t\!\bigl(\mathbf{w}(T;\,\mathbf{s})\bigr)\Big],
\end{equation}
which is unique by assumption~(A4a).
Note that $T_t^\star$ is a type-level constant (not a function of~$\mathbf{s}$).

By linearity of expectation,
\begin{equation}\label{eq:expect}
    \mathbb{E}\!\left[\sum_i w_i\, U_i\right]
    = \sum_i w_i\, \mu_i.
\end{equation}
By independence~(A1),
\begin{equation}\label{eq:var}
    \mathrm{Var}\!\left(\sum_i w_i\, U_i\right)
    = \sigma_t^2 \sum_i w_i^2.
\end{equation}
Thus,
\begin{equation}\label{eq:obj-expanded}
    \mathcal{R}_t(\mathbf{w})
    = \sum_i w_i\, \mu_i
    - \lambda\,\sigma_t^2 \sum_i w_i^2.
\end{equation}

\subsection{Effect of Temperature and Retrieval Uncertainty}\label{app:uncertainty}

The softmax temperature $T$ controls the concentration of $\mathbf{w}(T;\,\mathbf{s})$:
\begin{itemize}[nosep]
    \item As $T \downarrow 0$, $\mathbf{w}(T;\,\mathbf{s})$ becomes increasingly peaked (approaching one-hot), and $\sum_i w_i^2$ increases.
    \item As $T \uparrow \infty$, $\mathbf{w}(T;\,\mathbf{s})$ approaches the uniform distribution, and $\sum_i w_i^2$ decreases to $1/N$.
\end{itemize}
This reveals the core trade-off: decreasing $T$ increases both the alignment with high-$\mu_i$ passages (higher expected mean) and the variance of the fused signal (larger $\sum_i w_i^2$).
The variance penalty in $\mathcal{R}_t$ is scaled by $\sigma_t^2$, so:
\begin{itemize}[nosep]
    \item When $\sigma_t^2$ is small, the penalty is weak and the optimal temperature $T_t^\star$ is small (sharp fusion).
    \item When $\sigma_t^2$ is large, the penalty dominates and $T_t^\star$ shifts upward (flatter fusion).
\end{itemize}
More precisely, differentiating the expected objective (Eq.~\eqref{eq:obj-expanded}, with $\mathbf{w}=\mathbf{w}(T;\,\mathbf{s})$) with respect to $T$ and taking the expectation over $\mathbf{s}\mid t$ (justified by dominated convergence, since softmax derivatives are bounded) yields the first-order condition:
\begin{equation}\label{eq:foc-T}
    \mathbb{E}_{\mathbf{s}\mid t}\!\left[\sum_i \frac{\partial w_i}{\partial T}\,\mu_i\right]
    \;=\;
    2\lambda\,\sigma_t^2\;
    \mathbb{E}_{\mathbf{s}\mid t}\!\left[\sum_i w_i \frac{\partial w_i}{\partial T}\right].
\end{equation}
By the score--utility alignment assumption~(A3), the left-hand side is strictly decreasing in~$T$ near the optimum.
The right-hand side scales linearly with $\sigma_t^2$, and the derivative term is non-zero and of constant sign near $T_t^\star$ by~(A4b).
By the implicit function theorem applied to~\eqref{eq:foc-T}, $T_t^\star$ is a smooth function of $\sigma_t^2$.
When $\sigma_t^2$ increases, the right-hand side of~\eqref{eq:foc-T} increases at the old $T_t^\star$, violating the first-order condition; since the left-hand side is decreasing in $T$, the balance point must shift to a larger $T$.
Combined with uniqueness of the interior maximizer~(A4a), the balance point $T_t^\star$ is strictly increasing in $\sigma_t^2$.

\subsection{Impossibility of a Single Dataset-level Temperature}\label{app:impossibility}
By~(A2), there exist two types $A$ and $B$ with $\sigma_A^2 \neq \sigma_B^2$, each with positive probability.
By the strict monotonicity established in~\S\ref{app:uncertainty}, the type-level optimal temperatures (Eq.~\ref{eq:type-opt}) satisfy
\begin{equation}\label{eq:T-diff}
    T_A^\star \neq T_B^\star.
\end{equation}

Any dataset-level heuristic fixes a single temperature $T$ for all samples.
Since $T_A^\star \neq T_B^\star$ and both types occur with positive probability~(A2), any fixed $T$ differs from at least one type optimum.
By uniqueness of the interior maximizer~(A4a), using any $T \neq T_t^\star$ on type~$t$ yields a strict loss; that is, at least one of the following holds:
\begin{align}\label{eq:subopt}
    &\mathbb{E}_{\mathbf{s}\mid A}\!\Big[\mathcal{R}_A\!\bigl(\mathbf{w}(T;\,\mathbf{s})\bigr)\Big]
        < \mathbb{E}_{\mathbf{s}\mid A}\!\Big[\mathcal{R}_A\!\bigl(\mathbf{w}(T_A^\star;\,\mathbf{s})\bigr)\Big]
    \notag\\
    &\mathbb{E}_{\mathbf{s}\mid B}\!\Big[\mathcal{R}_B\!\bigl(\mathbf{w}(T;\,\mathbf{s})\bigr)\Big]
        < \mathbb{E}_{\mathbf{s}\mid B}\!\Big[\mathcal{R}_B\!\bigl(\mathbf{w}(T_B^\star;\,\mathbf{s})\bigr)\Big]
\end{align}
(or both), so the expected calibration regret~\eqref{eq:regret-def} is strictly positive.
In particular, this holds for $T_{\mathrm{global}}^\star$ (Eq.~\ref{eq:global-opt}), so $\mathrm{Reg}(T_{\mathrm{global}}^\star) > 0$.
This completes the proof that sample-level temperature adaptation is strictly beneficial under heteroscedastic retrieval uncertainty.
\qed

\subsection{Oracle Weights and Connection to FCPRAG}\label{app:oracle}
To make the architectural connection explicit, we derive the oracle weights in the $N=2$ case, relaxing the softmax parameterization to optimize over $w$ directly.
The resulting oracle provides a lower bound on what softmax-parameterized fusion can achieve; the first-order expansion below shows that FCPRAG approximates this oracle in the near-uniform regime.
Let $w_1 = w$ and $w_2 = 1-w$, with $\Delta\mu = \mu_1 - \mu_2 > 0$.
The objective~\eqref{eq:obj-expanded} becomes
\begin{equation}\label{eq:obj-n2}
    \mathcal{R}_t(w)
    = w\,\mu_1 + (1-w)\,\mu_2
    - \lambda\,\sigma_t^2\bigl[w^2 + (1-w)^2\bigr].
\end{equation}
Taking the derivative and setting it to zero:
\begin{equation}\label{eq:foc-w}
    \Delta\mu - 2\lambda\,\sigma_t^2\,(2w - 1) = 0
    \quad\Longrightarrow\quad
    w_t^\star = \frac{1}{2} + \frac{\Delta\mu}{4\lambda\,\sigma_t^2}.
\end{equation}
This oracle weight has the form \emph{uniform baseline $+$ utility-proportional deviation that shrinks with noise}, which directly motivates the $(g,\tau)$ gating mechanism: $g$ controls the deviation magnitude and $\tau$ controls the sharpness, together playing the role of $\Delta\mu/(4\lambda\,\sigma_t^2)$.
We can match this to FCPRAG's gated fusion (Eq.~\ref{eq:weights}).
For $N=2$, FCPRAG produces
\begin{equation}\label{eq:fcprag-n2}
    w_i^{\mathrm{FC}}
    = g\cdot\mathrm{softmax}(s_i/\tau)
    + (1-g)\cdot\tfrac{1}{2}.
\end{equation}
Under a first-order expansion around the near-uniform regime ($|s_1-s_2|/\tau \ll 1$; used here for interpretability, though the learned controller is not restricted to this regime):
\begin{equation}\label{eq:softmax-approx}
    \mathrm{softmax}(s_1/\tau)
    \;\approx\; \frac{1}{2} + \frac{s_1 - s_2}{4\tau},
\end{equation}
so
\begin{equation}\label{eq:fcprag-expand}
    w_1^{\mathrm{FC}}
    \;\approx\; \frac{1}{2}
    \;+\; \frac{g\,(s_1 - s_2)}{4\tau}.
\end{equation}
Comparing with the oracle~\eqref{eq:foc-w}: under score--utility alignment, $s_1-s_2$ is positively correlated with $\Delta\mu$ in expectation, so the oracle suggests setting $g/\tau$ to decrease with $\sigma_t^2$.
Note that $\lambda$ is a free parameter in the theoretical model with no direct architectural counterpart; the theory therefore provides qualitative guidance (the \emph{direction} of adaptation) rather than quantitative calibration, while the controller learns the appropriate scale from data.
In particular, the controller should increase $g/\tau$ (sharper, more confident fusion) when $\sigma_t^2$ is small and decrease it (conservative, near-uniform fusion) when $\sigma_t^2$ is large, which is precisely the adaptive behavior that FCPRAG learns from data.

\paragraph{Remark (observability).}
In practice, the Fusion Controller does not observe $\sigma_t^2$ directly;
it infers signals correlated with retrieval uncertainty from the query--passage interaction features (Eq.~\ref{eq:interaction}).
Proposition~\ref{prop:main} establishes that such adaptation is \emph{beneficial in principle};
whether the learned controller captures sufficient signal is an empirical question validated by our experiments.
The case studies in Appendix~\ref{app:cases} show that the learned $g$ and $\tau$ values vary meaningfully across queries with different evidence quality, suggesting the controller captures relevant uncertainty signals.

\subsection{Discussion}\label{app:theory-discussion}

This result formally motivates predicting fusion calibration (e.g., temperature and a gate interpolating with a uniform prior) at the sample level.
Such a mechanism can adapt to heteroscedastic retrieval uncertainty and avoid the inherent limitations of dataset-level fusion heuristics.
We note several simplifications in the theoretical model: the mean--variance objective is a tractable proxy for downstream QA metrics, the two-type mixture is a minimal construction (the argument extends to any finite number of types with distinct noise variances), and the $N=2$ oracle relaxes the softmax constraint.
Despite these simplifications, the qualitative prediction aligns with our empirical findings: FCPRAG's gains over PRAG are most pronounced on subtasks with heterogeneous evidence quality (Table~\ref{tab:main}, Figure~\ref{fig:retrieval_perturb}).
This is exactly the regime where the calibration regret of a global policy is largest: the retrieved set mixes strongly and weakly relevant passages, amplifying the cost of a fixed temperature.

\section{Additional Details}
\subsection{Use of Large Language Models}
\label{app:llm_usage}
Large language models were used for language polishing and proofreading to improve the clarity and readability of this manuscript.

\subsection{Extended Related Work}
\label{app:rw_extended}

\paragraph{In-context RAG and document aggregation.}
Classical RAG pipelines inject retrieved evidence by concatenating passages with the query or by aggregating document information inside the decoder~\cite{rag2020,dpr2020,fid2021,guu2020retrieval,borgeaud2022improving,izacard2022few,khandelwal2019generalization}. This design has been highly effective, but it also makes performance sensitive to context length, retrieval noise, and evidence conflicts, especially in multi-hop reasoning settings where useful and distracting passages may coexist~\cite{active_rag2023,selfrag2024,hotpotqa2018,ho2020multi,zhu2024fanoutqa,lu2024controlled}. Methods such as RePlug further improve how retrieved documents are used without updating the backbone LLM~\cite{shi2024replug}. Our setting differs because the retrieved evidence is fused in \emph{parameter space} rather than prompt space: each passage induces a candidate parametric update, and the core question becomes how strongly each update should influence the final merged adapter.

\paragraph{Expert routing and mixture-of-experts.}
\label{app:rw_moe}
Learning to select or weight components conditioned on an input has a long history in mixture-of-experts models.
Sparsely-gated MoE layers~\cite{shazeer2017outrageously} and efficient variants such as Switch Transformers~\cite{fedus2022switch} use routing to activate a subset of experts per input (or token) to improve capacity and efficiency; large-scale conditional computation systems such as GShard further highlight the benefits of input-dependent expert selection~\cite{lepikhin2020scaling}.
While our Fusion Controller is not an MoE layer inside the backbone LLM, it plays an analogous role at the \emph{evidence level}: it produces retrieval-conditioned weights over passage-specific parametric updates.
This difference in granularity matters: evidence-level fusion must decide not only which updates are helpful, but also how \emph{confidently} to deviate from uniform fusion under retrieval noise and evidence conflicts.

\paragraph{Calibration and re-weighting.}
\label{app:rw_calib}
Calibration methods such as Platt scaling and temperature scaling~\cite{platt1999probabilistic,guo2017calibration,niculescu2005predicting,kuleshov2018accurate} adjust predictive distributions to better reflect uncertainty.
In PRAG-family fusion, mapping controller scores to merge weights often requires careful temperature and blending choices.
\method{} integrates calibration into the fusion controller by learning a temperature and a mixing gate that blends learned weights with a uniform prior, yielding stable fusion without expensive global tuning.
This perspective aligns calibrated fusion with classical calibration: the goal is not only to rank evidence, but to control how \emph{confidently} the system deviates from uniform fusion when score signals are unreliable.
More broadly, reducing tuning and selection overhead is a recurring theme in RAG systems: hyperparameters that work well for one backbone or dataset can shift under changes in retrievers, context length, or evidence noise.
Our diagnostic analysis therefore treats calibration as an explicit design target rather than an afterthought.

\subsection{Inference Pseudocode}
\label{app:inference_pseudocode}

\begin{algorithm}[H]
\caption{Fusion-controlled parametric RAG inference (\method{}).}
\label{alg:fcp_inference}
\begin{algorithmic}[1]
\Require Query $q$; retrieved passages $\mathcal{P}=\{p_i\}_{i=1}^{N}$; passage adapters $\{\Delta_i\}_{i=1}^{N}$; Fusion Controller $f_\theta$; frozen backbone LLM $\mathcal{M}$; weighted adapter merger $\mathcal{M}_{\mathrm{WAM}}(\cdot)$.
\Ensure Generated answer $\hat{a}$.
\State Compute query embedding $\mathbf{e}_q$ and passage embeddings $\{\mathbf{e}_{p_i}\}$ using the sentence encoder.
\For{$i \gets 1$ to $N$}
  \State Build interaction feature $\mathbf{x}_i = [\mathbf{e}_q;\ \mathbf{e}_{p_i};\ \mathbf{e}_q \odot \mathbf{e}_{p_i};\ |\mathbf{e}_q-\mathbf{e}_{p_i}|]$.
\EndFor
\State Predict fusion scores $\{s_i\}$ and per-sample calibration $(g,\tau)$ using $f_\theta$.
\State $\pi \gets \mathrm{softmax}( \{s_i/\tau\}_{i=1}^{N} )$.
\For{$i \gets 1$ to $N$}
  \State $w_i \gets g \cdot \pi_i + (1-g)\cdot \frac{1}{N}$.
\EndFor
\For{$i \gets 1$ to $N$}
  \State $\tilde{w}_i \gets \frac{N\,w_i}{\sum_{j=1}^{N} w_j + \epsilon}$.
\EndFor
\State $\Delta^\star \gets \mathcal{M}_{\mathrm{WAM}}(\{\Delta_i\}, \{\tilde{w}_i\})$
\State Inject $\Delta^\star$ into $\mathcal{M}$ to obtain adapted model $\mathcal{M}_{\Delta^\star}$.
\If{\textbf{Combine} mode}
  \State Build the in-context prompt with $q$ and passages $\mathcal{P}$.
\Else
  \State Build the prompt with $q$ only (\textbf{Parametric-only}).
\EndIf
\State $\hat{a} \gets \textsc{Generate}(\mathcal{M}_{\Delta^\star}, \text{prompt})$.
\State \Return $\hat{a}$.
\end{algorithmic}
\end{algorithm}

\subsection{Reproducibility Details}
\label{app:implementation}

All experiments were conducted on 2× NVIDIA H100 GPUs.
We evaluate three instruction-tuned backbones: Llama-3.2-1B, Qwen2.5-1.5B, and Llama-3-8B~\cite{llama2023,2025_arXiv_Qwen2.5_Qwen2.5-Technical-Report}.
All compared methods share the same retriever and use top-$K=3$ passages per query, matching the default setting in PRAG-family evaluations.
We use \texttt{bge-base-en-v1.5} from SentenceTransformers for query and passage embeddings.
The Fusion Controller consists of two independent networks: a 3-layer scoring network (hidden dimensions: 2048, 1024) that predicts per-passage fusion scores, and a 2-layer calibration network (hidden dimension: 256) that predicts per-sample calibration parameters $(g, \tau)$.
Both networks use ReLU activations and dropout rate 0.1.
Training uses AdamW optimizer with learning rate $10^{-4}$, weight decay 0.01, batch size 32, and 10 epochs.
Following PRAG~\cite{prag2025}, we use LoRA rank $r=2$, scaling factor $\alpha=32$, and dropout $=0$.
For the multi-hop benchmarks, we follow a fixed-subset evaluation protocol per subtask and report subtask-level F1 together with the 2Wiki macro average.
For retrieval perturbation analysis, we corrupt one of the three retrieved passages via \textbf{replace-one} or \textbf{repeat-one}, and measure relative performance retention in the Parametric-only setting.

\paragraph{Runtime and overhead.}
The Fusion Controller operates on fixed-size query and passage embeddings and adds only a small MLP forward pass at inference time.
Unlike global fusion calibration via grid search, \method{} learns calibration parameters directly from data without backpropagating through the backbone LLM.

\subsection{Merge-aware Label Construction and Analysis}
\label{app:label_analysis}
\label{app:label_figure}

\noindent Figure~\ref{fig:labelgen} gives a schematic view of the leave-one-out label construction process described in Section~\ref{sec:labels}.

\begin{center}
  \maybeincludegraphics[width=\linewidth]{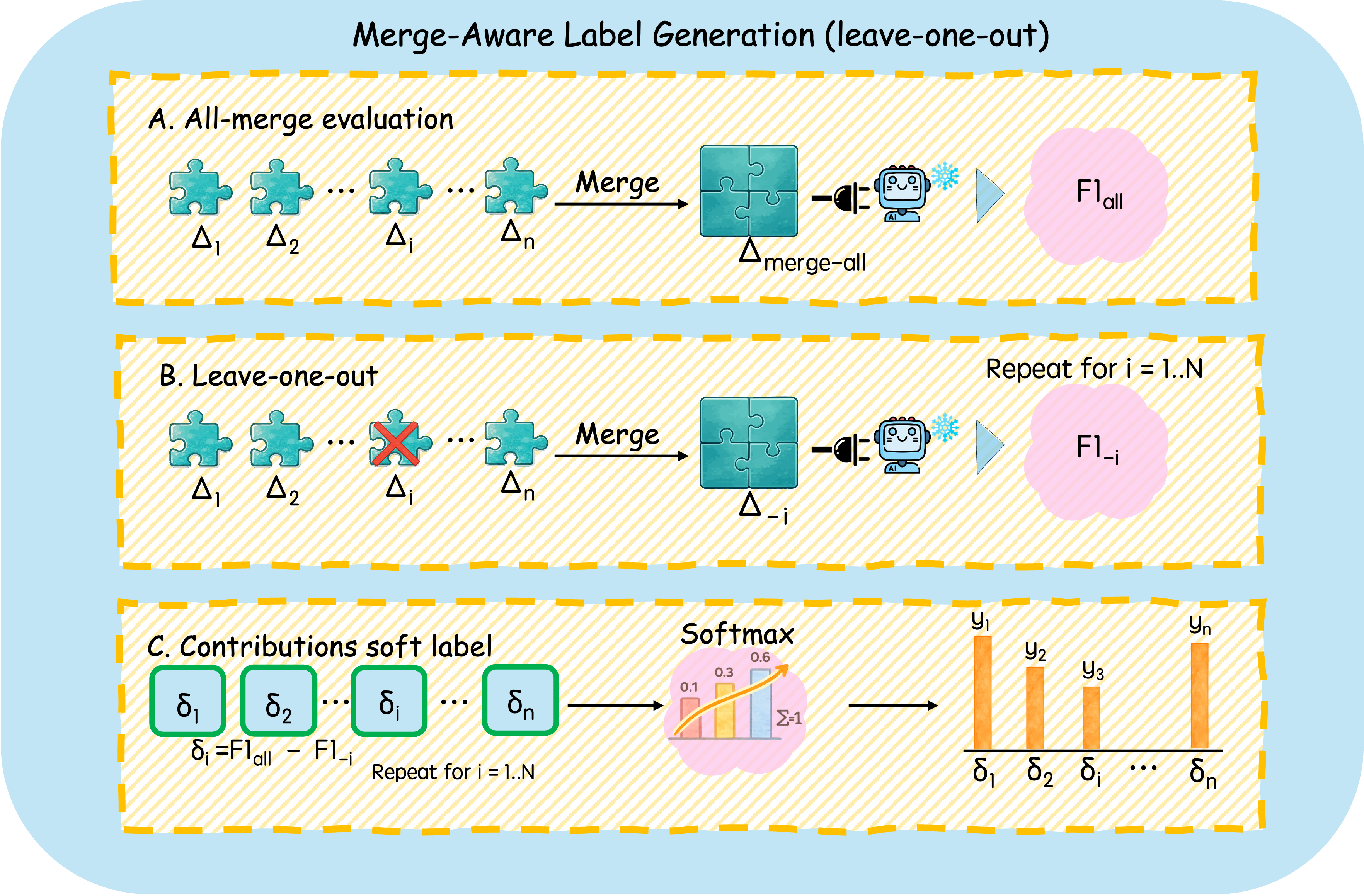}
  \captionof{figure}{Merge-aware label construction. For each sample, compute $F_1$ with all adapters merged and with each adapter removed; marginal contributions are softmax-normalized into a soft label distribution $y$ that supervises fusion-weight prediction. This procedure uses training data only.}
  \label{fig:labelgen}
\end{center}

\noindent To test whether merge-aware labels provide useful supervision beyond isolated adapter utility, we compare two label sources while keeping the scoring projector, backbone, and Parametric-only evaluation setting fixed.
Table~\ref{tab:label_source_ablation} shows that merge-aware labels improve macro F1 over single-adapter labels on HotpotQA.

\Needspace{10\baselineskip}
\begin{center}
\captionof{table}{Label-source ablation on HotpotQA (Qwen2.5-1.5B, Parametric-only).}
\label{tab:label_source_ablation}
\setlength{\tabcolsep}{5pt}
\renewcommand{\arraystretch}{1.05}
\resizebox{\linewidth}{!}{%
\begin{tabular}{lccc}
\toprule
\textbf{Label source} & \textbf{Bridge} & \textbf{Comparison} & \textbf{Macro} \\
\midrule
Single-adapter labels & 17.71 & 47.24 & 32.47 \\
Merge-aware labels & \textbf{17.95} & \textbf{49.87} & \textbf{33.91} \\
\midrule
$\Delta$ & +0.24 & +2.63 & +1.43 \\
\bottomrule
\end{tabular}%
}
\end{center}

\noindent We also compare isolated single-adapter utility rankings with the merge-aware leave-one-out rankings.
Table~\ref{tab:label_diagnostics} indicates that isolated adapter behavior is a weak proxy for the full merged ranking, even when the top contributor often agrees.
For $N$ retrieved adapters, label construction requires one all-merge evaluation and $N$ leave-one-out evaluations per training sample, i.e., $N+1$ merged evaluations; this cost is offline, uses training data only, and the resulting labels can be reused across controller variants.

\Needspace{9\baselineskip}
\begin{center}
\captionof{table}{Agreement between isolated single-adapter utility and merge-aware leave-one-out labels on HotpotQA (Qwen2.5-1.5B).}
\label{tab:label_diagnostics}
\setlength{\tabcolsep}{4.5pt}
\renewcommand{\arraystretch}{1.05}
\resizebox{\linewidth}{!}{%
\begin{tabular}{lccc}
\toprule
\textbf{Metric} & \textbf{Bridge} & \textbf{Comparison} & \textbf{Overall} \\
\midrule
Avg. Spearman $\rho$ & 0.0098 & 0.0157 & 0.0127 \\
Top-1 match (\%) & 76.00 & 58.50 & 67.25 \\
\bottomrule
\end{tabular}%
}
\end{center}

\subsection{Varying Number of Retrieved Passages}
\label{app:varying_k}

\noindent Table~\ref{tab:varying_k} reports a sensitivity check on 2WikiMultiHopQA with Qwen2.5-1.5B when the number of retrieved passages varies.
We include this analysis to test behavior beyond the default $K=3$ setting.
Parametric-only performance is not monotonic as $K$ increases, reflecting the difficulty of fusing more passage-specific updates without also injecting more noise.
In the Combine setting, however, \method{} improves over PRAG-Combine at $K=4$ and $K=5$.

\Needspace{11\baselineskip}
\begin{center}
\captionof{table}{Varying-$K$ results on 2WikiMultiHopQA with Qwen2.5-1.5B (F1, \%).}
\label{tab:varying_k}
\setlength{\tabcolsep}{5pt}
\renewcommand{\arraystretch}{1.05}
\resizebox{\linewidth}{!}{%
\begin{tabular}{lccc}
\toprule
\textbf{Method} & \textbf{$K=3$} & \textbf{$K=4$} & \textbf{$K=5$} \\
\midrule
Standard RAG & 25.97 & 28.58 & 28.82 \\
PRAG & 28.05 & 27.05 & 27.01 \\
PRAG-Combine & 26.04 & 31.12 & 33.90 \\
\method{} & \textbf{29.87} & 26.97 & 28.21 \\
\method{}-Combine & 27.38 & \textbf{33.63} & \textbf{35.08} \\
\bottomrule
\end{tabular}%
}
\end{center}

\subsection{Dataset Overview}
\begin{center}
\refstepcounter{table}
\textbf{Table~\thetable. Dataset overview and subtask structure.}
\label{tab:dataset_summary}

{\renewcommand{\arraystretch}{1.05}
\small
\resizebox{0.95\linewidth}{!}{%
\begin{tabular}{l l l}
\toprule
\textbf{Dataset} & \textbf{Domain / Focus} & \textbf{Subtasks} \\
\midrule
2WikiMultiHopQA & \makecell[l]{Wikipedia-based\\multi-hop QA} & \makecell[l]{Bridge-comparison; Comparison;\\Compositional; Inference} \\
HotpotQA & \makecell[l]{Wikipedia-based\\multi-hop QA} & Bridge; Comparison \\
PopQA & \makecell[l]{Entity-centric\\open-domain QA} & Single task \\
CWQ & \makecell[l]{Web-based\\complex QA} & Single task \\
\bottomrule
\end{tabular}}}
\end{center}

\subsection{Training-side Prompts}
\label{app:train_prompts}
This appendix lists the key prompt templates used to construct synthetic supervision and train passage-LoRA adapters offline.
Placeholders such as \texttt{\{passage\}} and \texttt{\{question\}} indicate runtime fields.

\paragraph{Passage rewrite (augmentation).}
\begin{PromptDiffBoxBlue}{Training prompt: passage rewrite}
Rewrite the following passage. While keeping the entities, proper nouns, and key details such as names, locations, and terminology intact, create a new version of the text that expresses the same ideas in a different way. Make sure the revised passage is distinct from the original one, but preserves the core meaning and relevant information.
{passage}
\end{PromptDiffBoxBlue}

\paragraph{QA-pair generation from a passage (augmentation).}
\begin{PromptDiffBoxBlue}{Training prompt: passage-to-QA generation}
I will provide a passage of text, and you need to generate three different questions based on the content of this passage. Each question should be answerable using the information provided in the passage. Additionally, please provide an appropriate answer for each question derived from the passage.
You need to generate the question and answer in the following format:
[
  {
    "question": "What is the capital of France?",
    "answer": "Paris",
    "full\_answer": "The capital of France is Paris."
  }
]
This list should have at least three elements. You only need to output this list in the above format.
Passage:
{passage}
\end{PromptDiffBoxBlue}

\paragraph{Passage-LoRA fine-tuning prompts (offline parameterization).}
Given QA pairs derived from passage $p_i$, we fine-tune a passage-specific LoRA adapter $\Delta_i$ using a mix of passage-conditioned and question-only prompts:
\begin{PromptDiffBoxBlue}{Training prompt: passage-conditioned and question-only fine-tuning}
(A) Passage-conditioned:
You should answer the question by referring to the knowledge provided below and integrating your own knowledge.
Passage 1: {passage}

Question: {question}

(B) Question-only (no passage in context):
You should answer the question by referring to the knowledge provided below and integrating your own knowledge.

Question: {question}
\end{PromptDiffBoxBlue}

\section{Case Studies}
\label{app:cases}
This section provides one real case per dataset from our main evaluation setting.
Each case reports the question, retrieved passages, rendered prompt structure, model output, and fusion signals.

\begin{PromptDiffBoxBlue}{2WikiMultiHopQA (Llama-3-8B): real case (combined inference)}
\textbf{Question.}\\
\texttt{Which film came out earlier, Morecambe Church Lads' Brigade At Drill or Little Funny Guy?}

\textbf{Retrieved passages (top-$K$).}\\
\texttt{Passage 1: ...Church Lads' and Church Girls' Brigade ... origins in 1891 ...}\\
\texttt{Passage 2: ...Church Lads' and Church Girls' Brigade ... historical notes ...}\\
\texttt{Passage 3: ...Boys' Life Brigade ... related youth-brigade timeline ...}

\textbf{Rendered prompt (combined setting).}\\
\begin{quote}\small\ttfamily
You should reference the knowledge provided below and combine it with your own knowledge to answer the question. Please follow the format of the example I provided above.\\
Here are some examples about how to answer the questions.\\
Question: When did the director of film Hypocrite (Film) die?\\
Answer: The film Hypocrite was directed by Miguel Morayta. Miguel Morayta died on 19 June 2013. So the answer is 19 June 2013.\\
Here are some reference.\\
Passage 1: ...Church Lads' and Church Girls' Brigade...\\
Passage 2: ...historical notes on Church Lads' Brigade...\\
Passage 3: ...Boys' Life Brigade and related organizations...\\
Let's think step by step. Answer the questions in the same format as above.\\
Question: Which film came out earlier, Morecambe Church Lads' Brigade At Drill or Little Funny Guy?
\end{quote}

\textbf{Model output and evaluation.}\\
\texttt{Generated text: ...Morecambe Church Lads' Brigade At Drill (1908) vs Little Funny Guy (1912)...}\\
\texttt{Eval answer: Morecambe Church Lads' Brigade At Drill}\\
\texttt{Gold answer: Morecambe Church Lads' Brigade at Drill}\\
\texttt{EM/F1: 1 / 1.0}

\textbf{Fusion signals (projector-enabled run).}\\
\texttt{scores: [0.5088, 0.5350, 0.5331]}\\
\texttt{g: 0.8357, tau: 0.0988}\\
\texttt{weights: [0.8644, 1.0767, 1.0588]}
\end{PromptDiffBoxBlue}

\begin{PromptDiffBoxBlue}{HotpotQA (Qwen2.5-1.5B): real case (combined inference)}
\textbf{Question.}\\
\texttt{Who has a wider scope of profession, José Echegaray or Graham Swift?}

\textbf{Retrieved passages (top-$K$).}\\
\texttt{Passage 1: María Catalina Irigoyen Echegaray ... Spanish Roman Catholic professed religious ...}\\
\texttt{Passage 2: ...Beatification record of María Catalina Irigoyen Echegaray ...}\\
\texttt{Passage 3: ...José Echegaray taught mathematics/hydraulics ... also served in government ...}

\textbf{Rendered prompt (combined setting).}\\
\begin{quote}\small\ttfamily
You should reference the knowledge provided below and combine it with your own knowledge to answer the question. Please follow the format of the example I provided above.\\
Here are some examples about how to answer the questions.\\
Question: Jeremy Theobald and Christopher Nolan share what profession?\\
Answer: Jeremy Theobald is an actor and producer. Christopher Nolan is a director, producer, and screenwriter. Therefore, they both share the profession of being a producer. So the answer is producer.\\
Here are some reference.\\
Passage 1: ...María Catalina Irigoyen Echegaray...\\
Passage 2: ...beatification context...\\
Passage 3: ...José Echegaray's teaching and political roles...\\
Let's think step by step. Answer the questions in the same format as above.\\
Question: Who has a wider scope of profession, José Echegaray or Graham Swift?
\end{quote}

\textbf{Model output and evaluation.}\\
\texttt{Generated text: ...José Echegaray was an engineer/politician/teacher, while Graham Swift is a novelist...}\\
\texttt{Eval answer: José Echegaray}\\
\texttt{Gold answer: José Echegaray}\\
\texttt{EM/F1: 1 / 1.0}

\textbf{Fusion signals (projector-enabled run).}\\
\texttt{scores: [0.4886, 0.4649, 0.4761]}\\
\texttt{g: 0.8178, tau: 0.0817}\\
\texttt{weights: [1.1235, 0.8864, 0.9901]}
\end{PromptDiffBoxBlue}

\begin{PromptDiffBoxBlue}{PopQA (Llama-3-8B): real case (combined inference)}
\textbf{Question.}\\
\texttt{What is Brigitte Bardot's occupation?}

\textbf{Retrieved passages (top-$K$).}\\
\texttt{Passage 1: ...Fondation Brigitte Bardot ... animal protection foundation ...}\\
\texttt{Passage 2: ...Brigitte Bardot Foundation ... created by French actress Brigitte Bardot ...}\\
\texttt{Passage 3: ...Brigitte Bardot ... French actress, singer, dancer, and fashion model ...}

\textbf{Rendered prompt (combined setting).}\\
\begin{quote}\small\ttfamily
You should reference the knowledge provided below and combine it with your own knowledge to answer the question. Please follow the format of the example I provided above.\\
Here are some examples about how to answer the questions.\\
Question: When did the director of film Hypocrite (Film) die?\\
Answer: The film Hypocrite was directed by Miguel Morayta. Miguel Morayta died on 19 June 2013. So the answer is 19 June 2013.\\
Here are some reference.\\
Passage 1: ...Fondation Brigitte Bardot...\\
Passage 2: ...French actress Brigitte Bardot...\\
Passage 3: ...Brigitte Bardot biography...\\
Let's think step by step. Answer the questions in the same format as above.\\
Question: What is Brigitte Bardot's occupation?
\end{quote}

\textbf{Model output and evaluation.}\\
\texttt{Generated text: ...Brigitte Bardot is a French actress, singer, dancer, and fashion model...}\\
\texttt{Eval answer: actress}\\
\texttt{Gold answer: [actor, actress, ...]}\\
\texttt{EM/F1: 1 / 1.0}

\textbf{Fusion signals (projector-enabled run).}\\
\texttt{scores: [0.5344, 0.4883, 0.5225]}\\
\texttt{g: 0.7373, tau: 0.1661}\\
\texttt{weights: [1.0854, 0.8860, 1.0286]}
\end{PromptDiffBoxBlue}

\begin{PromptDiffBoxBlue}{ComplexWebQuestions (Llama-3-8B): real case (combined inference)}
\textbf{Question.}\\
\texttt{What countries border the country whose capital is Sucre?}

\textbf{Retrieved passages (top-$K$).}\\
\texttt{Passage 1: Sucre ... constitutional capital of Bolivia...}\\
\texttt{Passage 2: What a Country! ... (distractor passage)}\\
\texttt{Passage 3: Capital Country ... New South Wales tourism region... (distractor)}

\textbf{Rendered prompt (combined setting).}\\
\begin{quote}\small\ttfamily
You should reference the knowledge provided below and combine it with your own knowledge to answer the question. Please follow the format of the example I provided above.\\
Here are some examples about how to answer the questions.\\
Question: When did the director of film Hypocrite (Film) die?\\
Answer: The film Hypocrite was directed by Miguel Morayta. Miguel Morayta died on 19 June 2013. So the answer is 19 June 2013.\\
Here are some reference.\\
Passage 1: Sucre... constitutional capital of Bolivia...\\
Passage 2: What a Country!...\\
Passage 3: Capital Country...\\
Let's think step by step. Answer the questions in the same format as above.\\
Question: What countries border the country whose capital is Sucre?
\end{quote}

\textbf{Model output and evaluation.}\\
\texttt{Generated text: ...countries bordering Bolivia are Brazil, Argentina, Paraguay, and Chile.}\\
\texttt{Eval answer: Brazil}\\
\texttt{Gold answer: [Brazil, Argentina, Paraguay, Chile, ...]}\\
\texttt{EM/F1: 1 / 1.0}

\textbf{Fusion signals (projector-enabled run).}\\
\texttt{scores: [0.5126, 0.5649, 0.5041]}\\
\texttt{g: 0.7724, tau: 0.1105}\\
\texttt{weights: [0.8840, 1.2808, 0.8352]}
\end{PromptDiffBoxBlue}

\end{document}